\documentclass{article}

\usepackage[preprint]{neurips_2025}

\usepackage[utf8]{inputenc}
\usepackage[T1]{fontenc}
\usepackage{amsmath,amssymb,amsthm}
\usepackage{mathtools}
\usepackage{graphicx}
\usepackage{wrapfig}
\usepackage{booktabs}
\usepackage{hyperref}
\usepackage{cleveref}

\newtheorem{remark}{Remark}
\newtheorem{conjecture}{Conjecture}

\DeclareMathOperator{\Cov}{Cov}

\newcommand{\reff}{\mathrm{rank}_{\mathrm{eff}}}

\title{Stream separation improves\\Bregman conditioning in transformers}

\author{
  J.\ Clayton Kerce \\
  Georgia Tech Research Institute \\
  \texttt{clayton.kerce@gtri.gatech.edu}
}

\date{}

\begin{document}

\maketitle

\begin{abstract}
Linear methods for steering transformer representations, including
probing, activation engineering, and concept erasure, implicitly assume
the geometry of representation space is Euclidean.
Park et al.~\citep{park2026information} showed that softmax induces a
curved Bregman geometry whose metric tensor is the Hessian of the
log-normalizer, $H(\lambda) = \Cov[\gamma \mid \lambda]$.  Ignoring
this curvature causes Euclidean steering to leak probability mass to
unintended tokens.  Their analysis applies at the output layer.
We measure this Hessian at intermediate layers in a controlled
$2 \times 2$ design crossing stream separation with per-layer
supervision (vocabulary decoding loss at each layer), all at matched
vocabulary and parameter count.
In standard single-stream transformers, $H$ is severely degenerate at
intermediate layers (effective rank 8 in 516 dimensions).
Stream separation improves conditioning by up to $22\times$ in
effective rank, even without auxiliary supervision.
Per-layer supervision helps, but less.
The cosine similarity between primal and dual concept directions
predicts per-layer steering effectiveness on downstream tasks, with
a threshold near $0.3$.
These results bear on the reliability of linear safety interventions,
which depend on the geometry being well-conditioned at the layer where
they are applied.
\end{abstract}
\footnotetext{This work was partially supported by DARPA Contract HR001125C0302.}

\section{Introduction}
\label{sec:intro}

Probing, activation engineering, and concept erasure are linear methods.
They identify a direction in representation space and add, remove, or
read off a signal along that direction~\citep{alain2017understanding,
turner2023activation, belrose2023leace}.  Each method assumes that
moving along a straight line in representation space produces a
predictable change in the output distribution.  This holds when the
geometry is flat.  It can fail when the geometry is curved.

The output layer of a transformer produces a probability distribution
over the vocabulary using the softmax function.  Softmax converts
logits $\lambda$ into a probability distribution
$p(y \mid \lambda) = \exp(\lambda^{\top}\gamma_y - A(\lambda))$,
where $A(\lambda) = \log \sum_y \exp(\lambda^{\top}\gamma_y)$ is the
log-normalizer, the function that controls how softmax distributes
probability mass across the vocabulary.  The Hessian of $A$ is
\begin{equation}\label{eq:hessian}
  H(\lambda) \;=\; \nabla^2 A(\lambda) \;=\; \Cov[\gamma \mid \lambda].
\end{equation}
Park et al.~\citep{park2026information} showed that $H$ is the metric
tensor of a dually flat Bregman geometry induced by softmax.  ``Dually
flat'' means the space has two natural coordinate systems, primal and
dual, related through $H$.  When $H$
is well-conditioned, primal directions approximate dual directions, and
linear steering works.  When $H$ is poorly conditioned, primal and dual
directions diverge, and Euclidean steering leaks probability mass to
unintended tokens.  Their analysis applies at the output layer.

We measure the conditioning of $H$ at every intermediate layer.  We
compare two interventions in a $2 \times 2$ factorial design: stream
separation (the CASCADE architecture~\citep{kerce2026dualstream}, which
maintains a frozen token-equivalent stream alongside a general
embedding stream to preserve symbolic representations throughout the
network) versus single-stream, crossed with per-layer supervision
(vocabulary decoding loss at the output of each feed-forward block) versus
no auxiliary loss.  All four models share the same vocabulary, parameter
count, and training data.

We find three things.
\begin{enumerate}
  \item In single-stream transformers without auxiliary loss, $H$ is
    severely degenerate at intermediate layers: effective rank 8 in a
    516-dimensional space.  Linear methods at these layers operate in
    2\% of the geometry.
  \item Stream separation improves conditioning more than the auxiliary
    loss.  CASCADE models without auxiliary loss have better-conditioned
    geometry than single-stream models with it.  Architecture, not
    training objective, is the primary determinant.
  \item The cosine similarity between primal and dual concept directions
    predicts per-layer steering effectiveness on downstream tasks.
    Below $0.3$, Euclidean steering is unreliable.  Above $0.4$, it
    works.  This gives practitioners a pre-deployment check.
\end{enumerate}

These results matter because linear safety interventions are
increasingly applied at intermediate layers.  A steering vector that
appears to work (the probe reads the intended signal) can silently
redistribute probability mass to unintended tokens if the geometry at
that layer is degenerate.  We provide a cheap diagnostic to detect this
failure mode and show that architecture choice determines whether it
arises.

\section{Background}
\label{sec:background}

Softmax induces a Bregman geometry on the space of logits.  The metric
tensor of this geometry is the Hessian $H(\lambda) = \Cov[\gamma \mid
\lambda]$ defined in \eqref{eq:hessian}.  Park et
al.~\citep{park2026information} proved that steering along the dual
geodesic (the natural gradient direction under $H$) is optimal at the
output layer: it moves probability mass to the target concept while
minimizing off-target disturbance (their Theorem~3).  We extend their
measurement to intermediate layers, where their optimality guarantee
does not directly apply.

The primal coordinates $\lambda$ are the logits.  The dual coordinates
$\eta = \nabla A(\lambda) = \mathbb{E}[\gamma \mid \lambda]$ are the
expected sufficient statistics, which for softmax are the token
probabilities $p(y \mid \lambda)$.  A small step in primal space maps
to a step in dual space through $d\eta = H(\lambda)\,d\lambda$.  A
primal direction is a straight line in logit space: the direction that
standard linear methods use.  A dual direction is the corresponding geodesic under the Bregman
metric: the direction that accounts for how softmax curves the space
and moves probability mass most efficiently.
When $H$ is well-conditioned, the two approximately agree, and linear
steering works.  When $H$ is ill-conditioned, some primal directions
are crushed to near-zero in the dual space while others are amplified.
Euclidean steering then leaks probability mass to unintended tokens.

We quantify conditioning with two measures.  The \emph{effective rank}
$\reff(H) = \exp\!\bigl(-\sum_i \hat\sigma_i \log \hat\sigma_i\bigr)$,
where $\hat\sigma_i = \sigma_i / \sum_j \sigma_j$ are the normalized
singular values, counts how many directions in $H$ carry meaningful
variance~\citep{roy2007effective}.  If the effective rank is 8 in a
516-dimensional space, then 98\% of the geometry is compressed into 8
directions, and a linear intervention that does not land in those
directions will have unpredictable effects on the output distribution.
The \emph{condition number}
$\kappa(H) = \sigma_{\max} / \sigma_{\min}$ measures the worst-case
ratio of amplification to compression in the primal-to-dual mapping:
how much $H$ distorts one direction relative to another.

\section{Experimental design}
\label{sec:design}

\paragraph{Models.}
We compare four transformer language models in a $2 \times 2$ factorial
design.  The first factor is architecture: \emph{stream separation}
versus \emph{single-stream}.  In the stream-separated (CASCADE)
architecture~\citep{kerce2026dualstream}, the token embedding stream
$x_t$ is frozen after initialization and a separate contextual stream
$x_e$ carries all learned computation; logits at any layer are computed
from $x_t + x_e^{(\ell)}$.  Because the token stream is always in the
coordinate system of the embedding matrix, logits at every layer are
computed in the same space as the output softmax.
In the single-stream architecture, a
single residual stream carries both token and contextual information,
as in a standard transformer.  The second factor is training objective:
an auxiliary cross-entropy loss at every intermediate layer (with weight
decaying linearly from $0.1$ at the first layer to $0.8$ at the final
layer) versus standard next-token loss only.  Architecture details are
in \cref{app:architecture}.

\paragraph{Controls.}
All four models use the GPT-2 tokenizer (50,257 tokens), have 45.4M
parameters, 6~layers, 6~heads, 516-dimensional embeddings, and gated
attention.  They are trained on the same data mixture for the same
number of steps.

\paragraph{Hessian measurement (Phase~1).}
For each model and each layer $\ell$, we extract pre-softmax logits at
the last token position across 480 contexts (30~batches of~16).  We
compute the sample covariance $\hat H^{(\ell)} = \Cov[\gamma \mid
\lambda]$ using a top-20K token approximation and report the effective
rank, condition number, and trace at each layer.

\paragraph{Steering comparison (Phase~2).}
We construct a binary concept direction from 15 gendered word pairs
(e.g., \emph{king}/\emph{queen}, \emph{boy}/\emph{girl}) using 40
gendered prompts.  From 10 starting contexts per layer, we steer
representations toward the target concept using both Euclidean addition
and dual (natural gradient) steering following the protocol of Park et
al.~\citep{park2026information}.  We measure off-target KL divergence
at the step where target concept probability first reaches 80\%.  The
difference in off-target KL between the two methods quantifies how much
the geometry penalizes Euclidean steering at each layer.  In practical
terms, this measures collateral damage: how much the rest of the output
distribution is disturbed while achieving the steering goal.

\section{The Hessian at intermediate layers}
\label{sec:hessian}

\begin{figure}[t]
\centering
\includegraphics[width=0.38\textwidth]{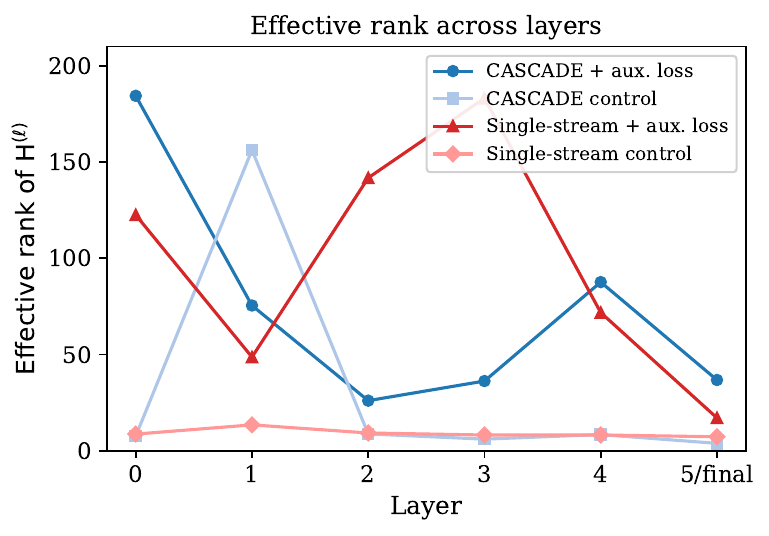}
\caption{Effective rank of $H^{(\ell)}$ across layers.  Single-stream
  control (pink diamonds) remains near 8.  Stream separation and
  auxiliary loss raise it by up to $22\times$.}
\label{fig:erank}
\end{figure}

The central finding is stark: in single-stream transformers, the
Hessian is severely degenerate at intermediate layers, while stream
separation largely fixes this.  \Cref{tab:erank} reports the effective
rank at each layer for all four models, \cref{tab:condition} reports
the condition number, and \cref{fig:erank} shows the effective rank
visually.

\begin{table}[t]
\centering\small
\caption{Effective rank of $H^{(\ell)}$ by layer.  All models have
  516~dimensions.  Higher is better.}
\label{tab:erank}
\begin{tabular}{@{}l*{4}{r}@{}}
\toprule
& \multicolumn{2}{c}{CASCADE} & \multicolumn{2}{c}{Single-stream} \\
\cmidrule(lr){2-3}\cmidrule(lr){4-5}
Layer & + aux.\ loss & control & + aux.\ loss & control \\
\midrule
0     & 184  & 8    & 123  & 9   \\
1     & 75   & 156  & 49   & 14  \\
2     & 26   & 9    & 142  & 9   \\
3     & 36   & 6    & 183  & 8   \\
4     & 88   & 8    & 72   & 8   \\
5     & 37   & 4    & 17   & 7   \\
final & 34   & 4    & 17   & 7   \\
\bottomrule
\end{tabular}
\end{table}

\begin{table}[t]
\centering\small
\caption{Condition number $\kappa(H^{(\ell)})$ by layer.  Lower is
  better.}
\label{tab:condition}
\begin{tabular}{@{}l*{4}{r}@{}}
\toprule
& \multicolumn{2}{c}{CASCADE} & \multicolumn{2}{c}{Single-stream} \\
\cmidrule(lr){2-3}\cmidrule(lr){4-5}
Layer & + aux.\ loss & control & + aux.\ loss & control \\
\midrule
0     & 1{,}297  & 14{,}701  & 2{,}185   & 49{,}392  \\
1     & 4{,}956  & 1{,}200   & 4{,}956   & 11{,}238  \\
3     & 5{,}672  & 19{,}768  & 1{,}217   & 20{,}142  \\
5     & 8{,}481  & 34{,}275  & 12{,}095  & 21{,}131  \\
final & 8{,}399  & 34{,}275  & 8{,}166   & 21{,}131  \\
\bottomrule
\end{tabular}
\end{table}

\paragraph{The baseline is degenerate.}
The single-stream control column of \cref{tab:erank} shows effective
rank between 7 and 14 at every layer.  The Hessian has 516 dimensions,
so only about 2\% of directions carry meaningful variance.  Any linear
method operating at these layers is working in a small subspace of the
Bregman geometry.

\paragraph{Stream separation fixes this.}
The CASCADE control column shows effective rank between 4 and 156, with
substantially higher values at layers 0 and 1.  More importantly, the
condition number (\cref{tab:condition}) drops by $2$--$7\times$
relative to single-stream control at layers 0 and 3--5.  This
improvement requires no auxiliary loss.  The only difference is the
frozen token stream.

The key comparison is CASCADE control versus single-stream with
auxiliary loss.  At deep layers (3--5 and final), CASCADE control has
condition numbers of $19{,}768$--$34{,}275$, while single-stream with
auxiliary loss reaches $1{,}217$--$12{,}095$.  The auxiliary loss
produces better condition numbers at these layers.  However, the
effective rank tells a different story: CASCADE control maintains higher
effective rank than single-stream with auxiliary loss at layers 4--5 and
final.  The two measures capture different aspects of conditioning, and
neither intervention dominates across all layers.

\paragraph{Auxiliary loss helps, but less.}
Adding auxiliary loss to the single-stream architecture raises effective
rank from 7--14 to 17--183, an improvement of $2$--$22\times$.  This
confirms that auxiliary supervision encourages better-conditioned
geometry.  But it does not close the gap with stream separation entirely.
At deep layers where distributions concentrate, the single-stream
auxiliary loss model still shows effective rank 17, while CASCADE with
auxiliary loss reaches 34--88.

\begin{figure}[b]
\centering
\includegraphics[width=0.38\textwidth]{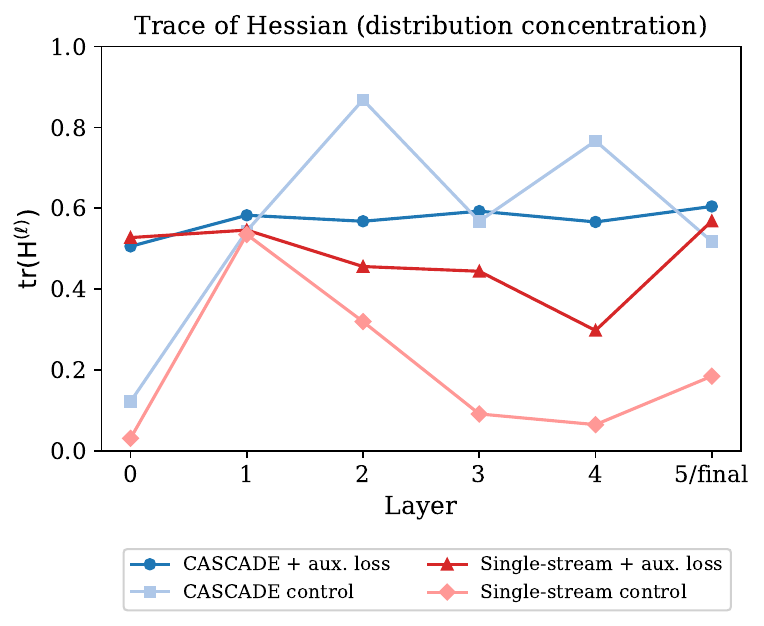}
\caption{Trace of $H^{(\ell)}$ across layers.  Single-stream control
  collapses at layers 3--4.  CASCADE maintains higher trace,
  preserving probability spread.}
\label{fig:trace}
\end{figure}

\paragraph{Distribution concentration.}
The trace of $H$ (total variance of the softmax distribution) explains
why deep layers behave differently from early layers.  When the trace
is near zero, the model is already confident about its prediction and
there is little probability mass left to steer.  In single-stream
models, the trace collapses from $0.5$ at layer~0 to $0.03$--$0.06$ at
layers 3--5: the softmax distribution concentrates on a few tokens
early in the network.  CASCADE models maintain traces of $0.4$--$0.7$
throughout.  This concentration affects the practical relevance of
conditioning: at layers where distributions are already peaked, both
good and bad geometry lead to similar steering outcomes, because there
is little probability mass left to misallocate.
\Cref{fig:trace} shows this pattern.

\section{Consequences for steering}
\label{sec:steering}

The conditioning differences in \cref{sec:hessian} predict that
Euclidean steering should fail where the Hessian is degenerate.  We
test this prediction.

\begin{table}[t]
\centering\small
\caption{KL advantage of dual over Euclidean steering
  ($\mathrm{KL}_{\text{Euclid}} - \mathrm{KL}_{\text{dual}}$ at 80\%
  target probability).  Positive values mean Euclidean steering causes unnecessary side
  effects that dual steering avoids.}
\label{tab:steering}
\begin{tabular}{@{}l*{4}{r}@{}}
\toprule
& \multicolumn{2}{c}{CASCADE} & \multicolumn{2}{c}{Single-stream} \\
\cmidrule(lr){2-3}\cmidrule(lr){4-5}
Layer & + aux.\ loss & control & + aux.\ loss & control \\
\midrule
0     & $+1.29$ & $-0.59$ & $+0.07$ & $+5.05$ \\
1     & $+1.51$ & $-0.05$ & $+0.53$ & $+3.19$ \\
2     & $-2.17$ & $+0.04$ & $-0.24$ & $+4.33$ \\
3     & $-0.56$ & $+0.15$ & $+0.16$ & $+0.59$ \\
4     & $-1.75$ & $+0.11$ & $-1.00$ & $-0.07$ \\
5     & $-0.76$ & $-0.31$ & $-0.47$ & $-0.15$ \\
final & $-0.47$ & $-0.31$ & $-0.44$ & $-0.15$ \\
\bottomrule
\end{tabular}
\end{table}

\paragraph{Where conditioning is bad, Euclidean steering fails.}
\Cref{tab:steering} shows the KL advantage of dual over Euclidean
steering at each layer.  In the single-stream control model, dual
steering outperforms Euclidean at layers 0--3 by $+0.6$ to $+5.1$~KL.
These are the layers where the Hessian is degenerate
(\cref{tab:erank}: effective rank 8--14).  Euclidean steering at these
layers leaks probability mass to unintended tokens.

In CASCADE models, the advantage largely vanishes.  The geometry is
well-conditioned enough that the simpler Euclidean method works.  In
CASCADE with auxiliary loss, the KL advantage is occasionally negative
(Euclidean slightly outperforms dual), consistent with near-flat
geometry where the overhead of computing the dual direction adds noise
without benefit.

At deep layers (4--final) in all models, both methods perform
similarly.  The absolute KL values are small ($<0.2$ for control,
$<1.0$ for auxiliary loss models).  As noted in \cref{sec:hessian},
distributions are concentrated at these layers, leaving little
probability mass to misallocate.

\paragraph{A cosine diagnostic predicts steering validity.}
The cosine similarity between the primal concept direction and the
corresponding dual direction tracks the regime boundary.  The intuition
is direct: when primal and dual point in roughly the same direction,
the curvature is mild and Euclidean steering is a good approximation.
When they diverge, the curvature is severe and Euclidean steering sends
probability mass in the wrong direction.
\Cref{tab:cosine} reports this quantity.

\begin{table}[t]
\centering\small
\caption{Cosine similarity between primal and dual concept directions.
  Values below $0.3$ correspond to layers where dual steering
  outperforms Euclidean in \cref{tab:steering}.}
\label{tab:cosine}
\begin{tabular}{@{}l*{4}{r}@{}}
\toprule
& \multicolumn{2}{c}{CASCADE} & \multicolumn{2}{c}{Single-stream} \\
\cmidrule(lr){2-3}\cmidrule(lr){4-5}
Layer & + aux.\ loss & control & + aux.\ loss & control \\
\midrule
0     & 0.58 & 0.42 & 0.18 & 0.09 \\
2     & 0.55 & 0.48 & 0.17 & 0.09 \\
4     & 0.62 & 0.58 & 0.41 & 0.30 \\
5     & 0.64 & 0.60 & 0.57 & 0.52 \\
final & 0.64 & 0.60 & 0.57 & 0.52 \\
\bottomrule
\end{tabular}
\end{table}

Where $\cos(\text{primal}, \text{dual}) < 0.3$, Euclidean steering is
unreliable: the primal direction does not approximate the dual geodesic,
and steering leaks probability mass.  Where
$\cos(\text{primal}, \text{dual}) > 0.4$, the two directions are
aligned enough that Euclidean steering works.  This pattern holds across
all four models and all layers.

\begin{wrapfigure}{r}{0.38\textwidth}
\centering
\vspace{-12pt}
\includegraphics[width=0.36\textwidth]{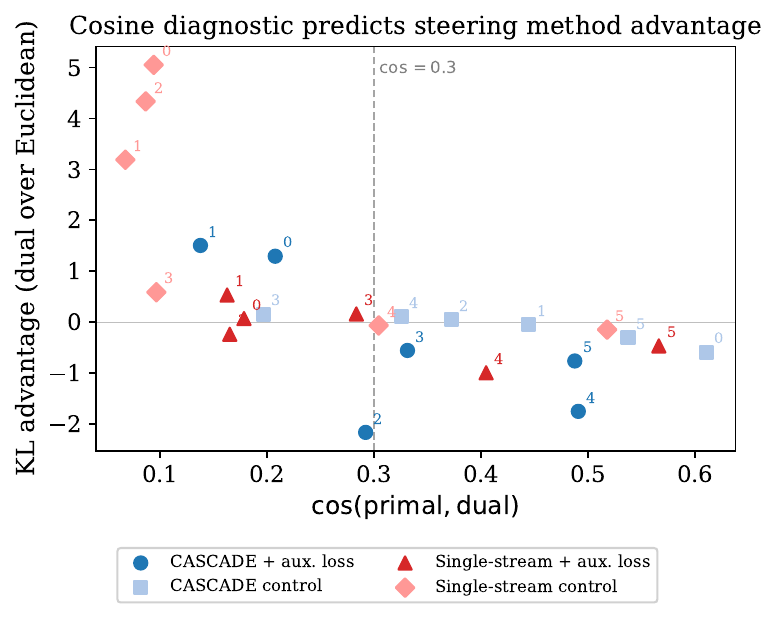}
\caption{Cosine diagnostic vs.\ KL advantage.  Each point is one
  (model, layer) pair.  Left of $\cos = 0.3$ (dashed), dual wins.
  Right of it, both methods perform similarly.}
\label{fig:cosine_kl}
\vspace{-10pt}
\end{wrapfigure}

The diagnostic is cheap to compute (one eigendecomposition of $H$) and
requires no steering experiment.  It can be evaluated before choosing
a steering method or layer.  \Cref{fig:cosine_kl} shows the
relationship between the cosine diagnostic and the KL advantage across
all models and layers.

\paragraph{Downstream validation.}
We validate the cosine diagnostic against direct vocabulary steering on
four tasks (coreference resolution, induction, recency, and
capitalization) using the same single-stream and CASCADE models with
layer$\times$scale sweeps across six step sizes.  Steering effectiveness
concentrates at layers 4--5, where $\cos > 0.3$ in the single-stream
model.  At layers 0--2, where $\cos < 0.1$ in the single-stream model,
effects are weaker and noisier, with occasional sign inconsistencies in
the coreference task.  CASCADE models show clean, sign-correct effects
at all layers, consistent with $\cos > 0.4$ throughout.  Scale sweeps
confirm linear, sign-correct scaling across all step sizes, indicating
that the dual steering advantage in \cref{tab:steering} is not an
artifact of step size choice.

\section{Related work}
\label{sec:related}

Our measurement extends the information-geometric analysis of Park et
al.~\citep{park2026information}, who characterized the Bregman geometry
of the output-layer softmax and proved that dual steering is optimal
there.  We ask whether their geometric structure is well-conditioned at
intermediate layers, and what architectural choices affect this.

The methods whose geometric assumptions we test include linear
probing~\citep{alain2017understanding}, activation
addition~\citep{turner2023activation}, representation
engineering~\citep{zou2023representation}, and concept
erasure~\citep{belrose2023leace}.  Each operates by identifying or
modifying a direction in representation space.  Each implicitly assumes
that straight lines in representation space correspond to meaningful
movements in output space.  Our results show that this assumption holds
at some layers and architectures but not others.

Information geometry in machine learning originates with
Amari's~\citep{amari2016information} natural gradient and the Fisher
information metric.  Frecon et al.~\citep{frecon2022bregman} studied
Bregman divergences as activation functions in neural networks.  Our
work uses the Bregman structure that softmax induces on its inputs,
following Park et al., rather than incorporating Bregman divergences
into the network architecture.

Several recent results connect architecture to interpretability.
Zhou et al.~\citep{zhou2025attention} identified a small number of attention
heads responsible for safety behavior, showing that interpretability
can be localized to architectural components.
Liu et al.~\citep{liu2025superposition} derived scaling laws for
superposition, linking representational capacity to the number of
features a network can encode.  The CASCADE architecture~\citep{kerce2026dualstream}
separates token and contextual streams to improve modularity.  Our
contribution is to show that this separation also improves Bregman
conditioning, providing a geometric account of why stream separation
aids interpretability.

\section{Discussion}
\label{sec:discussion}

We have established five facts.  First, the softmax Hessian at
intermediate layers of standard single-stream transformers is severely
degenerate, with effective rank roughly 2\% of the embedding dimension
(\cref{tab:erank}).  Second, stream separation improves this
conditioning without any change to the training objective
(\cref{tab:erank,tab:condition}).  Third, an auxiliary per-layer loss
also improves conditioning, but less than stream separation
(\cref{sec:hessian}).  Fourth, where the Hessian is degenerate,
Euclidean steering leaks probability mass to unintended tokens, and
dual steering reduces this leakage (\cref{tab:steering}).  Fifth, the
cosine similarity between primal and dual concept directions predicts
which layers are geometrically safe for linear intervention, with a
threshold near $0.3$ (\cref{tab:cosine}).  Together, these facts form
a chain: degenerate geometry causes steering failure, architecture
controls the geometry, and the cosine diagnostic detects the problem
before deployment.

\paragraph{Implications for safety and assurance.}
Linear interventions are increasingly used to enforce behavioral
constraints in deployed language models.  Activation engineering adds
or subtracts directions to promote refusal, reduce
toxicity, or suppress unwanted capabilities.  Concept erasure removes
directions to prevent models from representing sensitive attributes.
These methods assume that the geometry of representation space is
well-behaved: that a direction identified by probing corresponds to a
meaningful movement in the output distribution.

Our results show that this assumption is layer-dependent and
architecture-dependent.  In a single-stream transformer, linear
interventions at early layers operate in a 2\% subspace of the Bregman
geometry.  A safety-critical steering vector applied at such a layer
may not produce the intended behavioral change, or may produce it while
simultaneously disturbing the output distribution in uncontrolled ways.
The intervention appears to work in the Euclidean sense (the probe
reads off the intended signal) while failing in the geometric sense
(probability mass leaks to unintended tokens).

This is a verifiable failure mode, not a theoretical concern.  The
cosine diagnostic identifies the layers where linear interventions are
geometrically valid.  If $\cos(\text{primal}, \text{dual}) < 0.3$ at
a given layer, a linear intervention there cannot be trusted to produce
only the intended effect.  This check is cheap (one eigendecomposition)
and can be performed before deployment.

Stream separation provides an architectural path to reliable linear
intervention.  CASCADE models maintain $\cos > 0.4$ at all layers,
making Euclidean steering geometrically sound throughout the network.
This means that safety interventions applied at any layer will behave
as intended, without uncontrolled probability leakage.  The choice of
architecture is therefore not only a question of model capability or
training efficiency.  It is a question of whether the resulting geometry
supports the interpretability and control methods that will be applied
to the model after training.

\paragraph{Limitations.}
Our models are small (45.4M parameters, 6~layers).  Whether the
conditioning patterns persist at the scale of deployed language models
is an open question.  The steering evaluation uses gendered word pairs
as the primary concept; the downstream validation covers four tasks but
does not include safety-specific evaluations such as refusal or
toxicity.  The $0.3$ cosine threshold is empirical and may shift with
model scale or domain.

\paragraph{Future work.}
Whether the conditioning improvement arises from limiting cumulative
coordinate distortion across layers is a natural question.
\Cref{app:rigidity} states this as a conjecture.  A direct test would
require measuring the coordinate distortion (e.g., via the Jacobian of
the layer map) and correlating it with Hessian conditioning.  Extending the measurement to larger models and to
safety-specific steering tasks would strengthen the practical
relevance of the cosine diagnostic.

\bibliographystyle{plainnat}
\bibliography{references}

\appendix

\section{The rigidity conjecture}
\label{app:rigidity}

A standard transformer computes residual updates
$x^{(\ell+1)} = x^{(\ell)} + f_\ell(x^{(\ell)})$.  The map $f_\ell$
includes attention, a feed-forward network with nonlinear activations,
skip connections, and layer normalization.  Together these operations
can rotate, rescale, shear, and nonlinearly warp the coordinate system
from one layer to the next.  At an intermediate layer $\ell$, the
effective Bregman metric depends on how the remaining layers
$\ell+1, \ldots, L$ transform coordinates before reaching the output
softmax.  If the downstream layers distort the coordinate system
substantially, the primal direction at layer $\ell$ (a straight line
in layer-$\ell$ coordinates) will not correspond to a dual direction
in the output geometry.  The Hessian at layer $\ell$ will appear
poorly conditioned because the coordinate system is misaligned with
the output Bregman geometry.

The CASCADE architecture separates the representation into a frozen
token stream $x_t$ and a contextual stream $x_e$.  The logits at any
layer come from $x_t + x_e^{(\ell)}$.  Because $x_t$ is constant
across layers, it anchors the combined representation.  The contextual
stream $x_e$ is still subject to the full nonlinear layer map, but the
fixed $x_t$ component limits how far the total coordinate system can
drift from the output geometry.

\begin{conjecture}[Coordinate rigidity]\label{conj:rigidity}
  Stream separation improves Bregman conditioning at intermediate layers
  primarily by limiting the cumulative coordinate distortion between
  layers.  The frozen token stream anchors the coordinate system,
  reducing the misalignment between the layer-$\ell$ primal directions
  and the output-layer dual directions.
\end{conjecture}

\begin{remark}[Unit consistency]
An analogy to dimensional analysis in physics makes the conjecture more
concrete.  In physics, adding two quantities requires that they have
the same units.  The output softmax reads logits through the tied
embedding matrix $\mathbf{W}_E^\top$ (the same matrix that maps tokens
to embeddings, used in reverse to map representations back to
vocabulary scores), so the natural ``units'' of the output are
embedding-space coordinates.  In CASCADE, the token stream
$x_t = \mathbf{W}_E[t]$ is always in these units.  The contextual
stream $x_e^{(\ell)}$ must produce a vector addable to $x_t$ in the
same coordinate system, so unit consistency is maintained by
construction at every layer.  In a single-stream transformer, the
representation $x^{(\ell)}$ passes through LayerNorm (which rescales),
MLPs (which warp nonlinearly), and several composed layers.  Nothing
anchors it to the original embedding-space coordinates.  Applying
$\mathbf{W}_E^\top x^{(\ell)}$ at an intermediate layer computes a dot
product between embedding-space vectors and a vector that may no longer
be in embedding-space units.  The per-layer auxiliary loss adds
$\mathbf{W}_E^\top x^{(\ell)}$ at every layer, which acts as a soft
enforcement of unit consistency through gradient signal.  It helps, but
the architecture still permits the representation to drift between loss
computations.  CASCADE enforces unit consistency structurally.
\end{remark}

The auxiliary per-layer loss places a softmax head at every layer and
computes a cross-entropy loss against the target.  This adds gradient
signal encouraging each layer's representations to be useful for
prediction, but it does not constrain the coordinate system.  Each
layer's auxiliary head can learn its own projection to compensate for
coordinate drift.

Our results are consistent with the conjecture.  A direct test would
require measuring the coordinate distortion across layers (e.g., via
the Jacobian of the layer map) and correlating it with Hessian
conditioning.

\section{Architecture specification}
\label{app:architecture}

Full details of the dual-stream transformer are given
in~\citet{kerce2026dualstream}.  We summarize the elements relevant to
this paper.

\subsection{Dual-stream decomposition}

The residual stream at layer $\ell$ is decomposed as
\[
  \mathbf{x}^{(\ell)} = \mathbf{x}_t^{(\ell)} + \mathbf{x}_e^{(\ell)},
\]
where $\mathbf{x}_t^{(0)} = \mathbf{W}_E[t]$ is the token embedding
and $\mathbf{x}_e^{(0)} = \mathbf{0}$.

\subsection{CASCADE mode (frozen token stream)}

In CASCADE mode the token stream is never updated:
\begin{align}
  \mathbf{x}_t^{(\ell)} &= \mathbf{x}_t^{(0)}
    \quad \forall\, \ell, \\
  \mathbf{x}_e^{(\ell+1)} &= \mathbf{x}_e^{(\ell)}
    + \mathrm{Attn}^{(\ell)}(\mathbf{x}^{(\ell)})
    + \mathrm{FFN}^{(\ell)}(\mathbf{x}^{(\ell)}).
\end{align}
All updates accumulate in the contextual stream.  The combined
representation used for logits at any layer is
$\mathbf{x}^{(\ell)} = \mathbf{x}_t^{(0)} + \mathbf{x}_e^{(\ell)}$.

In the single-stream baseline, a single residual stream carries both
token and contextual information, as in a standard transformer.

\subsection{Per-layer supervision}

At each layer $\ell$, logits are computed through a shared language
modeling head with tied embeddings:
\[
  \mathbf{z}^{(\ell)}
    = \mathrm{LayerNorm}\!\bigl(\mathbf{x}^{(\ell)}\bigr)\,
      \mathbf{W}_E^\top.
\]
The training loss combines the final cross-entropy with weighted
auxiliary losses:
\[
  \mathcal{L}
    = \mathcal{L}_{\mathrm{CE}}(\mathbf{z}^{(L-1)}, y)
    + \lambda \sum_{\ell=0}^{L-2} w_\ell \,
      \mathcal{L}_{\mathrm{CE}}(\mathbf{z}^{(\ell)}, y),
\]
with $\lambda = 0.1$ and linear decay weights
$w_\ell = (\ell+1)/L$.

\subsection{Model variants}

\begin{center}
\begin{tabular}{@{}lccc@{}}
\toprule
Model & Stream sep. & Aux.\ loss & Params \\
\midrule
CASCADE + aux.\ loss   & \checkmark & \checkmark & 45.4M \\
CASCADE control        & \checkmark & ---        & 45.4M \\
Single-stream + aux.\ loss & ---   & \checkmark & 45.4M \\
Single-stream control  & ---        & ---        & 45.4M \\
\bottomrule
\end{tabular}
\end{center}

All models use 6 layers, 6 heads, 516-dimensional embeddings, gated
attention, the GPT-2 tokenizer (50,257 tokens), and are trained on the
same data mixture for the same number of steps.

\end{document}